\documentclass[conference]{IEEEtran}
\IEEEoverridecommandlockouts

\usepackage{cite}
\usepackage{amsmath,amssymb,amsfonts}
\usepackage{algorithmic}
\usepackage{graphicx}
\usepackage{textcomp}
\usepackage[dvipsnames]{xcolor}
\usepackage[hidelinks]{hyperref}
\def\BibTeX{{\rm B\kern-.05em{\sc i\kern-.025em b}\kern-.08em
    T\kern-.1667em\lower.7ex\hbox{E}\kern-.125emX}}

\usepackage[mathlines,switch]{lineno}

\usepackage{flushend}

\usepackage[strict]{changepage}

\usepackage{framed}

\definecolor{formalshadelight}{RGB}{242,242,242}
\definecolor{formalshadedark}{RGB}{166,166,166}

\usepackage{pifont}
\usepackage{comment}
\usepackage{todonotes}
\usepackage{tcolorbox}

\newboolean{CommentON}
\setboolean{CommentON}{false}

\usetikzlibrary{shapes.callouts, shadows}
\tikzset{author comment/.style={draw, fill=white, thick, drop shadow}}
\newcommand{\MyComment}[3]{%
	\ifthenelse{\boolean{CommentON}}{%
		\raisebox{-.5ex}
		{\tikz
			\node[x=1ex, y=1ex, inner sep=.5ex,
			rectangle callout,
			callout pointer width=.7ex,
			callout relative pointer={(1.5,-0)},
			author comment]
			{\footnotesize\textsf{#1}};}~%
		\textsf{[}\,\textcolor{#2}{#3}\,\textsf{]}
	}{} 
}

\newcommand{\rob}[1]{\MyComment{Rob}{orange}{#1}}
\newcommand{\mnote}[1]{\MyComment{Michelle}{teal}{#1}}

\newcommand{\highlight}[1]{\begin{tcolorbox}\textit{#1}\end{tcolorbox}}

\newcommand{\rqone}{Do AI algorithms differ in terms of energy consumption?}
\newcommand{\rqtwo}{Does modifying the dataset impact the energy efficiency of AI algorithms?}
\newcommand{\rqthree}{Can we improve the energy efficiency of AI algorithms through a data-centric approach without compromising their accuracy?}

\begin{document}

\title{Data-Centric Green AI \\An Exploratory Empirical Study}


\author{
\IEEEauthorblockN{Roberto Verdecchia\IEEEauthorrefmark{1}, Lu\'is Cruz\IEEEauthorrefmark{2},
June Sallou\IEEEauthorrefmark{3}, Michelle Lin\IEEEauthorrefmark{4}, James Wickenden\IEEEauthorrefmark{5}, Estelle Hotellier\IEEEauthorrefmark{6}}
\IEEEauthorblockA{\IEEEauthorrefmark{1}Vrije Universiteit Amsterdam, The Netherlands - \href{mailto:r.verdecchia@vu.nl}{r.verdecchia@vu.nl}}
\IEEEauthorblockA{\IEEEauthorrefmark{2}Delft University of Technology, The Netherlands - \href{mailto:l.cruz@tudelft.nl}{l.cruz@tudelft.nl}}
\IEEEauthorblockA{\IEEEauthorrefmark{3}Univ Rennes, France -  \href{mailto:june.benvegnu-sallou@irisa.fr}{june.benvegnu-sallou@irisa.fr}}
\IEEEauthorblockA{\IEEEauthorrefmark{4}McGill University, Canada -  \href{mailto:michelle.lin2@mail.mcgill.ca}{michelle.lin2@mail.mcgill.ca}}
\IEEEauthorblockA{\IEEEauthorrefmark{5}University of Bristol, United Kingdom -  \href{mailto:jw17943@bristol.ac.uk}{jw17943@bristol.ac.uk}}
\IEEEauthorblockA{\IEEEauthorrefmark{6}Inria, France -  \href{mailto:estelle.hotellier@inria.fr}{estelle.hotellier@inria.fr}}
}


\maketitle
\begin{abstract}
With the growing availability of large-scale datasets, and the popularization of affordable storage and computational capabilities, the energy consumed by AI is becoming a growing concern. To address this issue, in recent years, studies have focused on demonstrating how AI energy efficiency can be improved by tuning the model training strategy. Nevertheless, how modifications applied to datasets can impact the energy consumption of AI is still an open question. 

To fill this gap, in this exploratory study, we evaluate if data-centric approaches can be utilized to improve AI energy efficiency. To achieve our goal, we conduct an empirical experiment, executed by considering 6 different AI algorithms, a dataset comprising 5,574 data points, and two dataset modifications (number of data points and number of features). 

Our results show evidence that, by exclusively conducting modifications on datasets, energy consumption can be drastically reduced (up to 92.16\%), often at the cost of a negligible or even absent accuracy decline. As additional introductory results, we demonstrate how, by exclusively changing the algorithm used, energy savings up to two orders of magnitude can be achieved.

In conclusion, this exploratory investigation empirically demonstrates the importance of applying data-centric techniques to improve AI energy efficiency. Our results call for a research agenda that focuses on data-centric techniques, to further enable and democratize Green AI.

\end{abstract}


%

\section{Introduction}
 We live in the era of artificial intelligence (AI): new intelligent technologies are emerging every day to change people's lives. Many organizations identified the massive potential of using intelligent solutions to create business value. Hence, in the past years, the \textit{modus operandi} is collecting as much data as possible so that no opportunity is missed. Data science teams are constantly looking for problems where AI can be applied to existing data to train models that can provide more personalized and optimized solutions to their operations customers and operations~\cite{haakman2021ai}.

Nevertheless, the energy consumption of developing AI applications is starting to be a concern.
Previous studies observed that AI-related tasks are particularly energy-greedy~\cite{Strubell2019,Lacoste2019}.
In fact, since 2012, the amount of computing used for AI training has been doubling every 3.4 months~\cite{amodey2018ai}. Controversy has risen around particular machine learning models that have been estimated to consume the energy equivalent of a trans-American flight~\cite{Bender2021}.
Hence, a new subfield is emerging to make the development and application of AI technologies environmentally sustainable: \emph{Green AI}~\cite{Schwartz2020}.


On a related note, the current research practice of collecting massive amounts of data is not necessarily yielding better results. Being able to collect high-quality data is more important than collecting big data -- a trend coined as \emph{Data-centric AI}\footnote{Understanding Data-Centric AI: \url{https://landing.ai/data-centric-ai/}. Accessed 24th January 2022.}. Instead of creating learning techniques that squeeze every bit of performance, data-centric AI focuses on leveraging systematic, reliable, and efficient practices to collect high-quality data. 

Therefore, in this study, we conduct an exploratory empirical study on the intersection of Green AI and Data-centric AI. We investigate the potential impact of modifying datasets to improve the energy consumption of training AI models. In particular, we focus on machine learning, the branch of AI that deals with the automatic generation of models based on sample data -- machine learning and AI are used interchangeably throughout this paper.
In addition to investigating the energy impact of dataset modifications, we also analyze the inherent trade-offs between energy consumption and performance when reducing the size of the dataset -- either in the number of data points or features. Moreover, the analysis is performed in six state-of-the-art machine learning models applied in the detection of Spam messages.

Our results show that feature selection can reduce the energy consumption of model training up to 76\% while preserving the performance of the model. The improvement in energy efficiency is more impressive when reducing the number of data points: up to 92\% in the case of Random Forest. However, in this case, it is not cost-free: the trade-off between energy and performance needs to be considered. Finally, we also show that KNN tends to be the most energy-efficient algorithm, while ensemble classifiers tend to be the most energy greedy.


This paper provides insights to define the most relevant and energy-efficient modifications of datasets used during the development of the AI models while ensuring minimal accuracy loss. We argue that more research in Data-centric AI will help more practitioners in developing green AI models.
To the best of our knowledge, this is the first study to explore the potential of preprocessing data to reduce the energy consumption of AI. 

The entirety of our experimental scripts and results are made available with an open-source license, to enable the independent verification and replication of the results presented in this study: \url{https://github.com/GreenAIproject/ICT4S22}.

The remainder of this paper is structured as follows. Section~\ref{sec:relatedWork} presents the related work on the energy consumption of Artificial Intelligence models. Section~\ref{sec:studyDesign} details the overall approach and the study design. Section~\ref{sec:results} describes the results of the experimentation according to the different dataset modifications, and Section~\ref{sec:discussion} presents the related discussion. The threats to the validity of this study are thoroughly analyzed in Section~\ref{sec:threats}. Finally, Section~\ref{sec:conclusion} documents our conclusions and future work.


\section{Related Work} \label{sec:relatedWork}

Previous work has addressed the energy consumption of software systems across different domains, levels and ecosystems. There is ongoing research investigating how different frameworks~\cite{calero2021investigating}, data structures~\cite{oliveira2019recommending}, programming languages\cite{pereira2021ranking,georgiou2018your}, and so on, affect the energy consumption of software. The main outputs of the research in this field –– also known as Green Software -- aim at providing developers with informed advice on how to design, develop, and deploy their systems~\cite{anwar2020should,cruz2019catalog,cruz2019do,venters2018software,verdecchia2018empirical, verdecchia2017estimating}. Some works have also attempted at providing tools to help developers automatically improve the energy efficiency of their code\cite{ribeiro2021ecoandroid,cruz2019emaas}. Despite the numerous contributions in this field, only a handful of studies address the energy efficiency of AI-based systems~\cite{wyn}.

While numerous studies focus on utilizing AI to address sustainability concerns \cite{kloh, zhu}, only a few investigate how the sustainability of AI itself can be improved. Strubell~\textit{et al.} provide a clear landscape that motivates a research agenda in AI that considers their energy consumption~\cite{Strubell2019}. They pinpoint concrete cases of energy-intensive AI applications and compare the carbon emissions of training Natural Language Processing (NLP) models to ordinary daily tasks -- e.g., a car commute or air travel. Their results showcase that training a state-of-the-art NLP model can generate as much carbon as five cars during their entire lifespan (including fuel). Although our work also analyzes the energy consumption of training AI models, we aim at identifying trade-off decisions that can be generalized to other AI projects to reduce energy consumption.

In a similar direction, Schwartz~\textit{et al.} present the dichotomy between~\textit{Red AI} and~\textit{Green AI}. While traditional (Red) AI only aims to improve accuracy metrics, Green AI includes computational cost as a performance metric. Green AI favors the selection of algorithms that have comparable accuracy while consuming less energy. In their work, Schwartz~\textit{et al.} highlight the need for more research in the area of Green AI, showcasing the exponential growth of computational power required to train models over the past six years. Our work follows their call for a new research agenda in AI that brings energy consumption into the landscape of training an AI model. However, we take a step further by empirically investigating the potential of using data-centric over model-centric approaches to enable Green AI.

More research has been calling for a new research agenda in AI. Bender \textit{et al.}~\cite{Bender2021} provide a list of high-level recommendation to mitigate the unprecedented growth in the size of state-of-the-art NLP models. Recommendations include investing resources to curate datasets and reflecting on the potential risks entailed by models before developing them, to address AI sustainability. A different work reported concrete numbers on how the growth of AI is impacting the entire infrastructure of datacenters which need to grow in bandwidth, data storage, and power capacity~\cite{Wu2021}. While not focusing directly on AI sustainability, in other studies, researchers investigated the impact that utilizing smaller models~\cite{yang2021tuning} or down sampled datasets~\cite{zogaj2021doing} can have on accuracy.
Our study paves the way in directly addressing AI sustainability concerns by providing empirical evidence on how dataset modifications can be used to drastically save AI model training energy at a negligible accuracy loss.

Martin~\textit{et al.}~\cite{martin} focused on studying the energy consumption of a specific machine learning algorithm, namely the Very Fast Decision Tree (VFDT). The authors analyzed the energy consumption of VFDT at the function level, investigating how different parameters affect the energy consumption across all functions of the training algorithm. 
Their results demonstrate how function-level energy profiling can lead to improvements of up to 70\% in energy efficiency with minimal impact on the accuracy of the algorithm. 
In our research, we consider six different machine learning algorithms rather than a single one, as pinpointed in Section~\ref{sec:design}. Besides, we investigate for the first time if data-centric approaches can improve the energy efficiency of machine learning algorithms.

Previous work has studied the impact of machine learning algorithms in the context of mobile applications~\cite{mcintosh2018MLEnergy}. The authors compare eight mobile implementations of well-known training algorithms (e.g., k-Nearest Neighbor, Decision Trees, etc.) in terms of accuracy and energy consumption. In sum, the work shows that 1) energy consumption is often related to the algorithmic complexity of the algorithms, and 2) to achieve optimal energy efficiency practitioners ought to factor in application-specific variables -- e.g., whether the model needs to be regularly updated. Our work differentiates by 1) focusing on general-purpose implementations of machine learning algorithms rather than mobile-based ones and 2) providing a thorough analysis of the impact of the input data in the energy consumption of training a model.

Finally, a recent study analyzed the energy consumption of using different deep learning frameworks -- namely, PyTorch and TensorFlow~\cite{georgiou2022green}. Results suggest that TensorFlow achieves better energy performance at the training stage, while PyTorch is more energy-efficient at the inference stage. Our work differs by approaching energy efficiency from a data-centric perspective rather than a comparative analysis of different frameworks and libraries.

\section{Study Design and Execution} \label{sec:studyDesign}
In this section we document the empirical experiment executed for this study, in terms of goal (Section~\ref{sec:goal}), research questions (Section~\ref{sec:rq}), study subject (Section~\ref{sec:subject}), experimental procedure (Section~\ref{sec:procedure}), and data analysis (Section~\ref{sec:analysis}).

\subsection{Goal}
\label{sec:goal}
The aim of this research is to conduct an investigation into what influences the energy consumption of AI-based systems.
More formally, by utilizing the Goal-Question-Metric approach~\cite{gqm}, this objective can be described as follows:
\vspace{3pt}

\noindent\textit{%
    \textbf{Analyze} the energy consumption of model training\\
    \textbf{For the purpose of} identifying the impact\\
    \textbf{With respect to} dataset modifications\\
    \textbf{From the viewpoint of} software practitioners and researchers\\
    \textbf{In the context of} artificial intelligence.
}

\subsection{Research Questions}
\label{sec:rq}
In order to achieve our goal, we address the following three research questions (RQ):
\begin{LaTeXdescription}
    \item [$\mathbf{RQ_1}$] \textit{\rqone}
\end{LaTeXdescription}

By answering this introductory research question, we aim at understanding if AI algorithms impact differently the energy consumption of their underlying hardware, and in the affirmative case, the extent of this difference. The results gathered for this first research question allow us to gain sufficient knowledge on potential energy consumption difference of AI algorithms through which the following research questions, focusing on data-centric green AI, can be assessed.

\begin{LaTeXdescription}
    \item [$\mathbf{RQ_2}$] \textit{\rqtwo}\rob{Check for ``data properties" and refactor}
\end{LaTeXdescription}

While $RQ_1$ focuses on the potential difference in energy consumption of algorithms, with $RQ_2$ we explicitly focus on data-centric green AI, i.e., if modifications of the \textit{dataset} used by the algorithms can impact their energy consumption. Specifically, we split $RQ_2$ into two sub-RQs to study the potential impact of different facets of the dataset on the energy consumption of AI algorithms:

\begin{LaTeXdescription}
    \item[$\mathbf{RQ_{2.1}}$] \textit{Does the size of the dataset impact the energy consumption of AI algorithms?}
    \item[$\mathbf{RQ_{2.2}}$] \textit{Does the number of features impact the energy consumption of AI algorithms?}
\end{LaTeXdescription}
 
With $RQ_{2.1}$ we aim at understanding if utilizing only a portion of a dataset, instead of its entirety, can lead to a significant energy consumption difference of AI algorithms. Similarly, with $RQ_{2.2}$, we study if varying the number of features, i.e., the dimensionality of the dataset, can lead to a significant energy consumption variation.

While improving the energy efficiency of AI algorithms is at the core of our investigation, ensuring that energy efficiency improvements do not drastically deteriorate the effectiveness of AI algorithms, and hence defy their final purpose, is paramount. In order to systematically address this concern with our final research question, we investigate potential trade-offs between energy efficiency and algorithm accuracy (in terms of F1-score). This is expressed in $RQ_3$ as follows:

\begin{LaTeXdescription}
    \item[$\mathbf{RQ_3}$] \textit{\rqthree}
\end{LaTeXdescription}


\subsection{Experimental Subject}
\label{sec:subject}
In order to answer our RQs, we consider as experimental subject the \texttt{SMS Spam Collection} dataset~\cite{almeida2011contributions}. The~\texttt{SMS Spam Collection} is a dataset of labeled SMS messages collected for mobile phone spam research. The complete dataset is made publicly available at the University of California Irvine Machine Learning Repository\footnote{\url{https://archive.ics.uci.edu/ml/datasets/SMS+Spam+Collection}. Accessed 3rd January 2022.} and comprises 5,574 text message instances, labeled either as legit (``ham'' label, 4,827 instances) or spam (``spam'' label, 747 instances). The dataset is also made available \textit{via} the data science platform Kaggle\footnote{\url{https://www.kaggle.com/uciml/sms-spam-collection-dataset}. Accessed 3rd January 2022.}, where it was downloaded over 86,8K times, and used in more than 700 Jupiter notebook projects.

To preprocess our dataset, i.e., prepare the raw \texttt{SMS Spam} \texttt{Collection} data for the subsequent ``ham''/``spam'' classification, we make use of widely adopted standard techniques. Specifically, given that the \texttt{SMS Spam Collection} entails a text classification problem, we use a method involving term frequency–inverse document frequency (tf-idf), whereby words are tokenized based on their appearance in the dataset, and subsequently the term-frequency metric for each token is calculated. To execute the tokenization and term-frequency calculation, we utilize the standard implementation as provided in the Python package \texttt{scikit-learn}~1.0.\footnote{\url{https://scikit-learn.org/stable/modules/generated/sklearn.feature_extraction.text.TfidfTransformer.html}. Accessed 5th January 2022.} In total, the dataset includes 8169 features (i.e., 8168 token occurrence frequencies and a last feature corresponding to the length of the SMS messages).

In order to train and test our models, we utilize a 70\%/30\% train/test split. We do not allocate a portion of the dataset for validation purposes since, as further discussed in the threats to validity section (Section~\ref{sec:threats}), model optimization \textit{via} hyperparameter tuning falls outside the scope of this investigation.

\subsection{Experimental Procedure}
\label{sec:procedure}

\subsubsection{Experimental design}
\label{sec:design}
Our controlled empirical experiment is characterized by a set of Dependent Variables ($DV$) and Independent Variables ($IV$). We design the experiment as a set of \textit{treatments}, i.e., ``sub-experiments'' considering a specific combination of independent variable values.\footnote{While independent variable values vary among sub-experiments, the same set of dependent variables are collected for all sub-experiments.} For each sub-experiment, we exclusively vary one independent variable, while fixing all the other ones to a default level. This allows us to independently study the potential impact that each independent variable has on our dependent variables, while allowing us to adopt a straightforward and transparent research design. 

In addition, to answer our research questions, we are required to adopt a blocking factor, namely the factor \textit{AI algorithm} ($IV_1$). This entails that our sub-experiments are divided into different sets (or blocks), according to the specific utilized AI algorithm.

More specifically, in order to answer $RQ_1$, we consider the entire experimental dataset by fixing the number of data points ($IV_2$) and the number of features ($IV_3$) to their default level (i.e., 100\%), while exclusively varying the used AI algorithm ($IV_3$). This allows us to compare the energy consumption of AI algorithms ($DV_1$) in their ``default'' setting, i.e., without carrying out any \textit{ad hoc} manipulation of the original dataset.

To study the impact of dataset size ($RQ_{2.1}$) instead, we vary both the used AI algorithm ($IV_3$) and the number of data points ($IV_2$), by keeping the number of features ($IV_3$) to its default value. This allows to study the impact that the number of data points, i.e., the size of the dataset, has on the energy consumption of each algorithm ($DV_1$), while avoiding potential variation of experimental measurements due to different numbers of features.

Similarly, to answer ($RQ_{2.2}$), we vary the used AI algorithm~($IV_1$) and the number of features~($IV_3$), while fixing the number of used data points ($IV_2$) to its default value. This enables us to investigate the potential impact that the number of features has on the execution of AI algorithms ($DV_1$), while avoiding the potential impact on energy consumption due to variations of the number of data points.

Finally, to answer ($RQ_3$), we apply both experimental techniques employed to answer $RQ_2$, i.e., we vary AI algorithms~($IV_1$) and alternatively either the number of data points ($IV_2$) or number of features ($IV_3$), while fixing the other independent variable ($IV_3$ or $IV_2$) to its default value. This approach allows us to study independently the impact that the number of data point and the number of features have on accuracy ($DV_2$), while also enabling us to consider the data collected for $RQ_2$ to systematically answer this last RQ.

To ensure we gather statistically significant data, and to mitigate potential threats to internal validity, we repeat the execution of each sub-experiment 30 times. 
In addition, to mitigate the impact of potential confounding factors (e.g., an unnoticed execution of a background process affecting our energy measurements), rather than simply repeating sequentially the 30 executions of a sub-experiment, we shuffle the executions of sub-experiments uniformly at random.

An additional confounding factor may arise from the temperature of the utilized hardware.
To mitigate this threat, prior to the execution of our experiment, we perform a dummy CPU-intensive warm-up operation, carried out by calculating a Fibonacci sequence for approximately 5 seconds, and hence ensure that the hardware is not experiencing a ``cold boot'' when the first execution is run.

Finally, to avoid the potential influence of subsequent runs on our energy measurements, we introduce a sleep time equal to 5 seconds between each run, to allow the hardware to cool down, and execute all runs under the same initial hardware conditions.

\subsubsection{Experimental Variables}
\label{sec:variables}
Our experiment is characterized by a total of 3 independent variables and 2 dependent variables.

\textbf{Independent Variables (IVs).}
The independent variables of our experiment, i.e., the factors we adopt, and their corresponding values, are reported below. The default value of each independent variable (see Section~\ref{sec:design}) is distinguished with an under strike (except for the \textit{AI algorithm} independent variable, as it is our experimental blocking factor).  
\begin{itemize}
    \item \textbf{AI Algorithm ($\mathbf{IV_1}$):} \texttt{Support-Vector Machine}, \texttt{Decision Tree}, \texttt{Multinomial Naive Bayes}, \texttt{K-Nearest Neighbour}, \texttt{Random Forest}, \texttt{Adaptive Boost}, \texttt{Bagging Classifier}.
    \item \textbf{Number of data points ($\mathbf{IV_2}$):} \texttt{10\%}, \texttt{20\%}, \texttt{30\%}, \dots, \texttt{\underline{100\%}} of the total number of data points. To select data points, we adopt stratified sampling, and pick points of our population uniformly at random from each stratum (i.e., messages labeled as ``spam'' and ``ham'').
    \item \textbf{Number of features ($\mathbf{IV_3}$):} \texttt{10\%}, \texttt{20\%}, \texttt{30\%}, \dots, \texttt{\underline{100\%}} of the total number of features. To select features, we adopt the Chi-Square Test (Chi2)~\cite{liu1995chi2}, to ensure that only the most relevant features are considered for each level.
\end{itemize}

The set of AI algorithms ($IV_1$) was chosen by considering the most popular ones. We use the implementation provided in the Python library \texttt{scikit-learn}\footnote{\url{https://scikit-learn.org/stable/whats_new/v1.0.html\#version-1-0-0}. Accessed 3rd January 2022.}, which was used to implement the algorithms for this study. The discretization step size of the number of data points and features (10\%) was instead adopted to ensure sufficient granularity of results, while guaranteeing an \textit{a-priori} feasible number of experimental runs.

\textbf{Dependent Variables (DVs).} In terms of metrics used to answer our research questions, i.e., our observed dependent variables, we consider the following ones in our experiment:
\begin{itemize}
    \item \textbf{Energy consumption ($\mathbf{DV_1}$):} the energy consumed by the hardware on which the AI algorithms are executed, measured in Joules (J);
    \item \textbf{F1-score ($\mathbf{DV_2}$):} Overall accuracy measure of the model, defined as $F1=2* \frac{P*R}{P+R}$, where $P$ is the model precision, and $R$ the model recall.
\end{itemize}

The energy consumption ($DV_1$), measured during the execution of AI algorithms, is the dependent variable used to answer $RQ_1$ and $RQ_2$. The F1-score ($DV_2$) is instead adopted to answer $RQ_3$. The F1-score is chosen over precision ($P$) and recall ($R$) metrics, as it allows us to gain an encompassing summary overview of the overall accuracy of AI algorithms, while overcoming potential representation problems due to the uneven distribution of labels present in the dataset used (see Section~\ref{sec:subject}). 

\subsubsection{Experimental Setting} \label{sec:exp_setting}
All sub-experiments are run on a 
machine equipped with a 2.4GHz Quad-Core i5 processor and 16 GB 2133 MHz LPDDR3 of memory.
The entirety of the experiment and data analysis is implemented in Python 3.10.\footnote{\url{https://www.python.org/downloads/release/python-3100/}. Accessed 3rd January 2022.}
In order to measure energy consumption ($DV_1$), we leverage \texttt{codecarbon}\footnote{\url{https://github.com/mlco2/codecarbon}. Accessed 3rd January 2022.}, a Python package allowing to estimate the energy consumption of code running on Intel and AMD CPU processors.
All the AI algorithms ($IV_1$) follow the implementation as provided in the Python package \texttt{scikit-learn} 1.0, and use the standard hyperparameters as defined in the library. 

In total, by considering the combination of independent variables and sub-experiment repetitions, 3.6K experimental runs are executed to gather data to answer our research questions.

\subsection{Data Analysis}
\label{sec:analysis}
In this section, we report the data analysis procedure that we adopt to derive our results from the gathered data. 

As a preliminary step, in order to assess if the energy consumption ($DV_1$) data we collected is normally distributed, we carry out a visual normality assessment by means of quantile-quantile (Q-Q) plot, followed by a Shapiro-Wilk normality test. From the inspection of the generated Q-Q plot, and the Shapiro-Wilk test result ($W$=0.52 and p-value=2.2e-16), we can confidently conclude that the data collected is not normally distributed. Hence, for each sub-experiment, we sample the data gathered in the run reporting the median energy consumption value. Subsequently, in order to evaluate if a correlation exists between our dependent and independent variables, we leverage the calculation of the one-tailed Spearman's rank correlation coefficient ($\rho$). We adopt Spearman's~$\rho$ as it provides a non-parametric measure, and can be used to calculate the potential correlation between our ordinal ($IV_1$-$IV_3$) and continuous variables ($DV_1$ and $DV_2$). Finally, to provide further insights into our results, we calculate percentage changes to summarize the difference in energy consumption and F1-scores between different algorithms, number of data points, and number of features.

\section{Results} \label{sec:results}
In this section, we report the results of our empirical experimentation according to the research questions guiding this study
(see Section~\ref{sec:rq}). 

\subsection{Results \texorpdfstring{$RQ_1$}{RQ1}: Energy Consumption Variability of AI Algorithms}
\begin{figure}[hbpt!]
\includegraphics[width=\columnwidth]{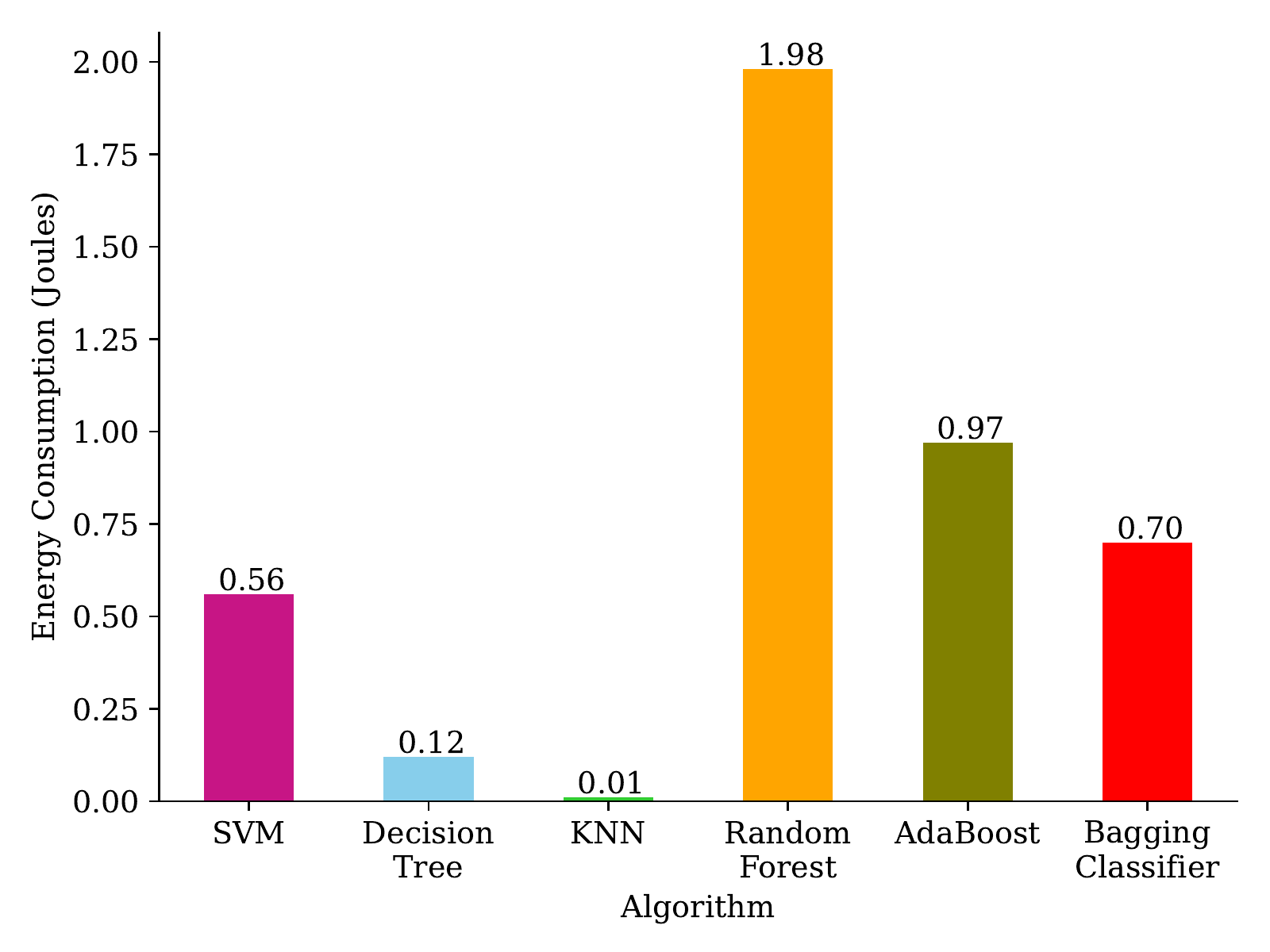}
\caption{Median Energy Consumption of AI Algorithms}
\label{fig:rq1}
\end{figure}

With our first research question, we aim at investigating the potential difference between the energy consumption of AI algorithms. 
An overview of the median consumption of each AI algorithm, as measured in our empirical experiment, is depicted in Figure~\ref{fig:rq1}.

By inspecting Figure~\ref{fig:rq1}, we can immediately notice that the energy consumption drastically varies among AI algorithms. More specifically, Random Forest results to be the most energy greedy algorithm, with a median energy consumption of 1.98 Joules per run, followed by AdaBoost, which nevertheless resulted to consume less than half (48.9\%) of the energy required by Random Forest. The most energy efficient algorithm results to be KNN, which reports a median energy consumption of 0.01 Joules, followed by Decision Tree, which requires 0.12 Joules. By considering minimum and maximum variation values, we note that energy consumption varies between algorithms from a minimum decrease of 20\% (Bagging Classifier - SVM) up to a 99.49\% decrease in energy consumption (Random Forest - KNN).

\subsection{Results~\texorpdfstring{$RQ_2$}{RQ2}: Impact of dataset modifications on energy consumption}
\begin{figure*}[hpt!]
\includegraphics[width=\textwidth]{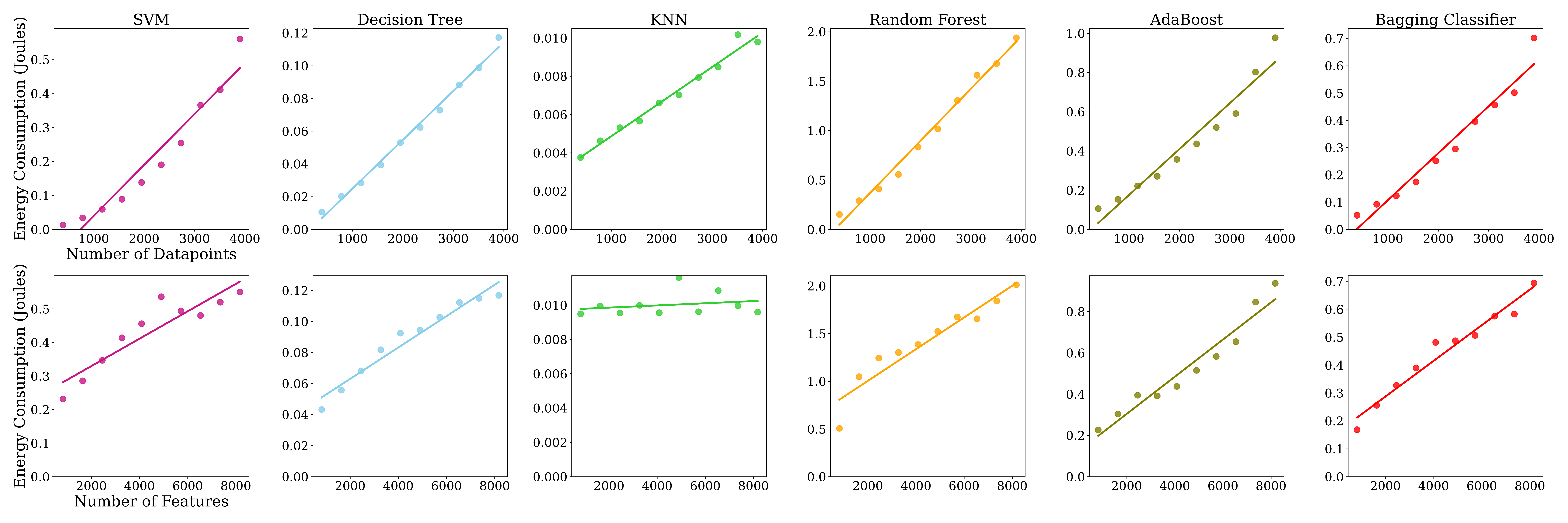}
\caption{$\mathbf{RQ_2}$ results. Small multiples showing the energy consumption of the different algorithms when using different data points (first row) and when using different numbers of features (second row).}
\label{fig:rq2}
\end{figure*}

With $RQ_2$, we aim at investigating if dataset modifications, and more specifically the number of data points ($RQ_{2.1}$) and the number of features ($RQ_{2.2}$), may have an impact on the energy consumed by AI algorithms. An overview of the results we collected for $RQ_2$ are depicted in Figure~\ref{fig:rq2}, and are further described below.

\subsubsection{Results~\texorpdfstring{$RQ_{2.1}$}{RQ2.1}: Impact of the number of data points on energy consumption}
\label{sec:rq2.1}
The first row of diagrams reported in Figure~\ref{fig:rq2} depicts the median energy consumption of each algorithm at a varying number of data points (reported on the $x$-axis). As we can intuitively notice from the linear regression lines reported in the plots, the energy consumption appears to be correlated with the number of data points. This observation is confirmed by the Spearman's rank correlation coefficient values reported in Table~\ref{tab:spearman}. By considering the $\rho$ values reported in Table~\ref{tab:spearman}, we note that there is a definitive positive correlation between the number of data points and the energy consumption, of either strong nature (i.e., $0.70\leq \rho \leq 0.89$ for KNN, Random Forest, Bagging Classifier) or very strong nature (i.e., $\rho \geq 0.90$ for SVM, Decision Tree, and AdaBoost). The corresponding p-values showcase that the identified correlations are with very low probability~due~to~chance. 

By considering the energy reduction achieved by using fewer data points, we notice that this independent variable ($IV_2$) influences the considered AI algorithms differently, and can lead to a maximum energy reduction ranging from 61.72\% (KNN) up to 92.16\% (Random Forest).

\begin{table}[]
\centering
\caption{Correlation analysis between energy consumption ($DV_1$) and number of data point ($IV_2$) or number of features ($IV_3$)}
\begin{tabular}{llcr}
\hline
 Algorithm ($IV_1$)     &  Indep. Variable   & $\rho$ &      p-value \\
\hline
 SVM                & Num. data points ($IV_2$) &  0.95  & 7.16e-151 \\
 Decision Tree      & Num. data points ($IV_2$) &  0.92   & 9.58e-120 \\
 KNN                & Num. data points ($IV_2$) &  0.80  & 3.24e-68  \\
 Random Forest      & Num. data points ($IV_2$) &  0.87  & 4.25e-95  \\
 AdaBoost           & Num. data points ($IV_2$) &  0.91  & 6.64e-115 \\
 Bagging Classifier & Num. data points ($IV_2$) &  0.87  & 3.07e-92  \\
 \hline
 SVM                & Num. features ($IV_3$) &  0.69  & 3.09e-43  \\
 Decision Tree      & Num. features ($IV_3$) &  0.75  & 2.29e-56  \\
 KNN                & Num. features ($IV_3$) &  \textbf{0.04} & \textbf{0.54 }    \\
 Random Forest      & Num. features ($IV_3$) &  0.64  & 2.02e-36  \\
 AdaBoost           & Num. features ($IV_3$) &  0.79   & 6.01e-66  \\
 Bagging Classifier & Num. features ($IV_3$) &  0.76  & 4.54e-58  \\
\hline
\end{tabular}
\label{tab:spearman}
\end{table}

\subsubsection{Results~\texorpdfstring{$RQ_{2.2}$}{RQ2.2}: Impact of the number of features on energy consumption}
The second row of diagrams reported in Figure~\ref{fig:rq2} depicts the energy consumption for each algorithm at a varying number of features (reported on the $x$-axis). From the distribution of median energy consumption values, and the linear regression lines, the number of features and the energy consumption appear to be correlated for most algorithms.
The relationship is confirmed by the Spearman's rank correlation coefficient values $\rho$ reported in Table~\ref{tab:spearman}. In comparison with the number of data points (see Section~\ref{sec:rq2.1}), the number of features results to possess an overall weaker positive correlation with the energy consumption, while still being either strongly (i.e., $0.70\leq \rho \leq 0.89$, for Decision Tree, AdaBoost, Bagging Classifier) or moderately correlated ($0.40\leq \rho \leq 0.69$ for SVM and Random Forest). Interestingly, varying the number of features does not noticeably affect the energy consumed by KNN, by showcasing only a very weak correlation ($0.0\leq \rho \leq 0.19$), which was with high probability dictated by chance (p-value=0.54). 

For all algorithms other than KNN, the energy reduction obtained by varying the number of features results to be lower than the one obtainable by varying the number of data points, while still being appreciable. As for the number of data points, varying the number of features affects differently the energy consumption of the considered AI algorithms. Interestingly, for KNN, lowering the number of features leads in numerous cases to a higher energy consumption w.r.t. the case of using all features. In addition, the best energy efficiency achieved by KNN by lowering the number of features results to be only a 0.92\% decrease. In comparison, the algorithm which showcases the highest energy efficiency by varying the number of features is AdaBoost, which achieves up to a 75.8\% energy reduction when compared to its baseline. 

\subsection{Results~\texorpdfstring{$RQ_3$}{RQ3}: Trade-offs between energy consumption and accuracy}
\begin{figure*}[hbpt!]
\includegraphics[width=\textwidth]{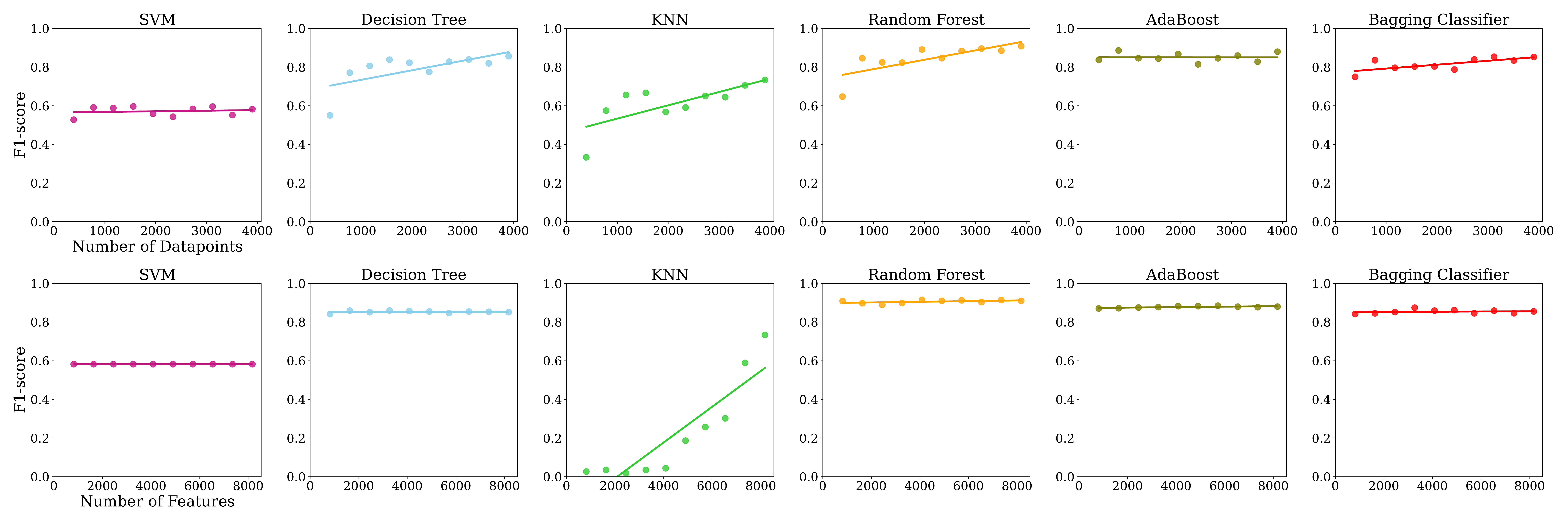}
\caption{RQ3 results. Small multiples showing the F1-score of the different algorithms when using different data points (first row) and when using different numbers of features (second row).}
\label{fig:rq3}
\end{figure*}

With $RQ_3$, we aim at investigating if potential trade-offs between AI energy efficiency and accuracy are possible. An overview of the accuracy results, in terms of F1-score collected \textit{via} our empirical experiment, is reported in Figure~\ref{fig:rq3}. As described in the figure, both by varying the number of data points ($IV_2$, first row of Figure~\ref{fig:rq3}) or the number of features ($IV_3$, second row of Figure~\ref{fig:rq3}) we generally do not observe a notable F1-score decrease (reported on the $y$-axis, Figure~\ref{fig:rq3}), with both numbers of data points and features not being correlated to F1-scores.

More detailed insights into the correlation analysis are provided by the Spearman’s rank correlation coefficient values $\rho$ reported in Table~\ref{tab:spearman_f1}. From the $\rho$ values reported in the table, we notice that, when considering the number of data points as independent variable, most algorithms report only a very weak correlation with F1-scores (for SVM and AdaBoost) or a moderate correlation (for KNN and Bagging Classifier). The only exceptions are the algorithms Decision Tree and Random Forest, both reporting a strong correlation between number of data points and F1-score. The relative p-values indicate that such correlation is statistically significant, i.e., w.h.p. not due to chance. 

When considering the correlation between number of features and F1-score, a different picture emerges. In fact, from the $\rho$ values reported in Table~\ref{tab:spearman_f1}, we can observe for most algorithms that the number of features is correlated to the F1-score either \textit{via} a very weak correlation (for Decision Tree and Bagging Classifier), or a moderate one (for Random Forest and AdaBoost). The $\rho$ value is not definable (N.D.) for SVM, as no variation is observed in F1-score values, i.e., the covariance between number of features and F1-score is zero. Interestingly, KNN is the only algorithm which reports a $\rho$ value indicating a very strong correlation between number of features and F1-score. By inspecting the relative p-value, we can conclude that such correlation is statistically significant. 



\begin{table}[]
\centering
\caption{Correlation analysis between F1-score ($DV_2$) and number of data points ($IV_2$) or number of features ($IV_3$)}
\begin{tabular}{llrr}
\hline
 Algorithm ($IV_1$)     &  Indep. Variable   & $\rho$ &      p-value \\
\hline
 SVM                & no\_datapoints     &     -0.018 &     0.960 \\
 Decision Tree      & no\_datapoints     &      \textbf{0.733} &     \textbf{0.016} \\
 KNN                & no\_datapoints     &      0.661 &     0.038 \\
 Random Forest      & no\_datapoints     &      \textbf{0.855} &     \textbf{0.002} \\
 AdaBoost           & no\_datapoints     &     -0.006 &     0.987 \\
 Bagging Classifier & no\_datapoints     &      0.661 &     0.038 \\
 \hline
 SVM                & no\_features       &    N.D.     &   N.D.     \\
 Decision Tree      & no\_features       &     -0.042 &     0.907 \\
 KNN                & no\_features       &      \textbf{0.954} &     \textbf{1.788e-05} \\
 Random Forest      & no\_features       &      0.541 &     0.106 \\
 AdaBoost           & no\_features       &      0.585 &     0.075 \\
 Bagging Classifier & no\_features       &      0.316 &     0.374 \\
\hline
\end{tabular}
\label{tab:spearman_f1}
\end{table}

\section{Discussion} \label{sec:discussion}

This empirical experiment provides exploratory evidence of the potential of using data preprocessing techniques to reduce the energy consumption of AI.
Below, we answer each research question by analyzing the results of our experimentation. Several conclusions are drawn with regard to the energy consumption of AI models and the impact of the input dataset modifications.

\subsection{\rqone~(\texorpdfstring{$RQ_1$}{RQ1})}

Yes, different training algorithms yield considerably different energy footprints. 
The algorithm with the least energy consumption is KNN, using almost $200\times$ less energy than Random Forest. However, that does not necessarily mean that KNN should always be chosen, as we prove later with research questions $RQ_2$ and $RQ_3$. Nevertheless, the energy consumption data collected with our experiment showcases the importance of logging such information.
Practitioners resort to different performance metrics when selecting and tuning models.
In agreement with previous work~\cite{Schwartz2020}, we argue that practitioners will consider different models when they are aware of these differences w.r.t. energy consumption. Hence, selecting a machine learning model should be a trade-off analysis encompassing not only accuracy metrics but also energy metrics.

Random Forest, AdaBoost, and Bagging classifiers were the most energy greedy algorithms. This is somehow expected since they all belong to a class of algorithms known as \textit{ensembles}, which combine the results of training multiple classifiers (a.k.a. weak learners) using slightly different parameters or training datasets. In other words, the energy consumption of ensembles is equivalent to training multiple models: it is affected by the number of weak learners being used internally and their individual energy consumption.  


To make energy metrics available to machine learning practitioners, we need better and more accessible ways of measuring energy consumption. As seen in this study, collecting energy consumption is not a trivial task. We need simple techniques to approximate energy consumption. Although this is out of the scope of this study, other studies suggest looking at duration, CPU usage, or the number of floating point operations~\cite{Strubell2019,radovanovic2021carbonaware,garciamartin}. Ideally, metrics could estimate energy consumption before even training the models -- i.e., by using static analysis approaches. 

The experimental nature of machine learning can also magnify the energy consumption reported in this paper. Practitioners have to retrain their models several times before converging to a final model. Previous studies have suggested this to increase energy consumption by a factor of roughly 2000$\times{}$: Strubell~\textit{et al.}~\cite{Strubell2019} show that, while training one of their natural language processing models has an electricity cost of \$5, the electricity cost of performing the full R\&D required to develop that model is estimated to be \$9,870.  Hence, small improvements in energy efficiency in the early stages of the pipeline can lead to large savings in the long run.

\vspace{2pt}
\highlight{\textbf{Main findings $\mathbf{RQ_1}$ (Algorithm Energy Consumption Comparison):}
Different algorithms yield completely different energy footprints. The difference goes up to a 99.49\% energy consumption decrease, with KNN being the most energy-efficient algorithm, and Random Forest Energy the least energy-efficient one. We argue that easy-to-use energy metrics are quintessential when selecting the best model for a machine learning project.}

\subsection{\rqtwo~(\texorpdfstring{$RQ_2$}{RQ2})}



Yes, except for KNN, all algorithms yield less energy consumption when we reduce the dimensionality of the dataset. In other words, there is a positive correlation between the size of the dataset and the measured energy consumption: using fewer data leads to more energy efficiency. Improvements go up to 92\% when reducing the number of rows and up to 76\% when reducing the number of features. Hence, instead of collecting the biggest amount of data, we must aim for smaller but meaningful datasets.

Our results demonstrate the importance of adding \mbox{data-centric} Green AI as a key topic in the research agenda of AI development. 
For example, recent work on core set extraction (i.e., extracting the smallest subset that keeps the key dataset properties) and dataset distillation have shown promising results in AI applications~\cite{nguyen2021dataset,zhao2021dataset,borsos2020coresets}. We argue that such strategies have a potential on Green AI that has been overlooked in previous research.

Although this work focuses on the dimensionality of data, it paves the way to study other properties of the input data. For example, one can expect that the data types used when loading the data into memory lead to different energy footprints when training the model. However, this kind of control is not fully supported by AI libraries/frameworks. For example, the widely utilized library adopted for this study, \texttt{scikit-learn}, automatically converts all data to floating-point with 64 bits (as of version 0.24.2). Users have no way of intervening in this data transformation. It is not yet known whether using data types with unnecessarily high precision can lead to unnecessary energy costs. We argue that this is a missed opportunity in AI libraries.  This is amplified if we consider IoT systems, where small devices are used to collect, process, and transmit data. Depending on the use case, these devices may operate with different precision levels -- e.g., 16 or 32 bits.
Developers of AI libraries ought to reconsider some design choices, to give back control to their users and enabling further energy efficiency opportunities.

Based on our results, we foresee potential in studying data-centric techniques to democratize Green AI. Several AI-leading organizations are aiming to be carbon-free by 2030. This requires massive investments in infrastructure and is far from being a realistic norm for the rest of the AI industry. For example, previous work on Green AI bring awareness to the importance of using energy-efficient hardware, datacenters in locations with better access to clean energy, etc.~\cite{verdecchia2021green,Lacoste2019}. While such measures are important, they might be inaccessible to most practitioners and organizations that operate on tight budgets.
Our results show that, with very simple techniques available to any AI practitioner, one can effectively reduce the carbon footprint of developing AI models. 

It is important to note that there is an energy consumption overhead when we manipulate the number of rows and columns.
We did not factor in this overhead as we focus on studying the impact of the dataset shape and not preprocessing techniques. This is an important detail since machine learning techniques such as cross-validation or parameter tuning will require training the model multiple times with the same preprocessed data. Nevertheless, assuming that each development cycle executes model training and data preprocessing exactly once, we observed an overhead revolving around 5\% on average. Moreover, most machine learning projects already resort to data reshaping methods for other purposes beyond energy efficiency. More research is needed to help practitioners define trade-offs, but our results show evidence that, as a rule of thumb, row sampling and feature selection should always be considered.

Another remark relates to the fact that improvements in the efficiency of AI are being followed by a massive increase in the usage of AI-based systems -- the so-called rebound effect. This is also referred to in another fields as the Jevons paradox~\cite{alcott2005jevons}, i.e, there is a correlation between the usage of natural resources and the improvements in the efficiency of a given technology.
In particular, improvements in the energy efficiency of AI are often targeted at leveraging more AI models in contexts where energy resources are prohibitively scarce -- for example, AI-based apps for smartphone devices. In these contexts, improvements on energy efficiency aim at delivering more AI systems, failing to reduce the overall carbon footprint of AI. We argue that our study is less prone to this rebound effect, as it provides meaningful advice that can be used by an AI project and lead to immediate savings on energy consumption. Nonetheless, we call for more research in Green AI that investigates the rebound effects in this context.

\vspace{2pt}
\highlight{\textbf{Main findings $\mathbf{RQ_2}$ (Impact of dataset modifications on energy consumption)}: Extracting smaller datasets poses a great opportunity to reduce the energy consumption of our machine learning models.
Improvements in energy efficiency go up to 92\% when we reduce the size of the dataset.
Our results call for more research on data-centric techniques to enable and democratise Green AI and for AI frameworks to give more control over how data is manipulated. }

\subsection{\rqthree~(\texorpdfstring{$RQ_3$}{RQ3})}
Yes, one can use data preprocessing techniques to improve energy efficiency without compromising the accuracy of the models. When reducing the number of features~($IV_3$) in the dataset, the trained models perform the same, in terms of accuracy, while consuming less energy. In other words, we prove that using more data does not always mean better models, while it leads to energy efficiency improvements.

The exception to this rule is KNN, which shows a strong positive correlation between the F1-score and the number of features in the dataset. This observation gets more interesting if we combine it with the results from $RQ_2$, where we observe that reducing the feature space does not yield any significant energy improvement. Reducing the number of features not only does not influence the energy efficiency of KNN, but also hinders the performance score.

Using SVM did not produce a model with a F1-score above 0.6, denoting an overall very low accuracy of the generated model relative to the other algorithms examined. 
A parameter tuning strategy to accommodate the imbalance and sparsity of our data could have yielded models with higher accuracy. However, we did not delve into optimization strategies (see Section~\ref{sec:threats} for more details), since we wanted to compare the energy consumption between different algorithms in a fair and intuitive way. Hence, the energy improvements observed for SVM in $RQ_2$ require further scrutiny before drawing generalizable conclusions.

Random Forest consistently yields the best performance. Despite being the most energy-greedy algorithm, it trained the most accurate model. Other algorithms, AdaBoost, Bagging Classifier, and Decision Tree follow close behind, showing competitive results. Nevertheless, it is not possible to fairly say which algorithm is the best. Different algorithms may work better with different problems. Once again, our observations reiterate that energy metrics provide useful information when selecting the best machine learning model, and are quintessential for AI development.


When it comes to the number of data points~($IV_2$), AdaBoost and Bagging Classifier yield no correlation between the size of the dataset and F1-score, meaning that the energy improvements shown earlier in $RQ_2$ do not bring any cost in the performance of the models.

For the remaining algorithms, the results are not as unanimous. Random Forest, Decision Tree, and KNN yield a positive correlation between cardinality and F1-score. This means that, despite the benefits in energy consumption, reducing the number of data points might prompt losses in the accuracy of these models. Nevertheless, this is worth considering because the models still showcase reasonable performance: with Decision Tree and Random Forest, F1-score drops less than 0.05 when we use at least 20\% of the original dataset. Depending on the use case, such a drop in F1-score might be appropriate.
\vspace{2pt}

\highlight{
\textbf{Main findings $\mathbf{RQ_3}$ (Trade-offs between AI energy consumption and accuracy):}
In the vast majority of cases, decreasing the number of data points / features drastically reduces energy consumption, while implying only a negligible accuracy deterioration (e.g., by reducing features, Random Forest can achieve a maximum of 74.81\% energy reduction at the cost of only a 0.06\%  F1-score reduction).
However, this observation does not hold for all algorithms. For example, feature selection when using KNN has almost no impact on energy consumption, while considerably reducing its model performance (with a maximum of 0.92\% energy reduction, associated to a~98.05\%~F1-score~loss).}

\mnote{@luis -- this line confuses me and may confuse the reader. Feature selection is typically not used in KNN ie. sklearn doesn't have a param for this and typically in order to do informal feature selection before training you would need to put another algorithm "on top" of it and this would involve dim. reduction (which is for the purpose of hyper param tuning -- which we did not do). I'm not quite sure how KNN feature selection was performed in this case -  I'll have to double check the scripts}


\section{Threats to Validity}
\label{sec:threats}
In this section, we discuss the threats to validity of our study, by following the categorization provided by Wholin~\textit{et~al.}~\cite{wohlin2012experimentation}.

\subsection{Conclusion Validity}
A threat to conclusion validity in our study could be constituted by low statistical power of the tests used to answer our RQs. To mitigate this threat, we systematically collected and analyzed data by following a process defined \textit{a priori}. By considering the combination of factor, treatments, and reruns, a total of 3.6K data samples were used to answer our RQ.

As a threat to the reliability of measures, unknown tasks running in the background during the execution of our experiment may have acted as confounding factors, hence influencing our energy measurements. To mitigate this threat, prior to the experiment execution, we ensured that only the piece of software necessary to run the experiment was running and/or able to be executed. In addition, each experiment was repeated 30 times, by shuffling the executions of sub-experiments uniformly at random, to avoid that potential confounding factors could influence only a specific set of sub-experiments.

\subsection{Internal Validity}

A threat to internal validity, related to history, could be constituted in our experiment by the influence that executing subsequent runs could have had on our measurements (e.g., due to hardware increasing temperature). To mitigate this threat, each experimental run was preceded by a 5-second sleep operation, to allow all runs to be executed under identical hardware conditions. Similarly, a warm-up operation was performed to ensure the first run was executed under the same conditions as the subsequent ones (see Section~\ref{sec:design}).


\vspace{4pt}
\subsection{Construct Validity}
A mono-method bias may have affected the results of our study, as we utilized a single metric to measure energy consumption ($DV_1$,~Section~\ref{sec:variables}) and calculate algorithm accuracy ($DV_2$,~Section~\ref{sec:variables}).

Regarding energy consumption, we do not deem that adopting only energy consumption as dependent variable constitutes per se a prominent threat in our experiment. In addition, utilizing exclusively energy consumption measurements is a common practice in the field of software energy efficiency~\cite{hindle2014greenminer,verdecchia2018code,feitosa2017investigating, verdecchia2018empirical}. However, relying on  a specific tool to estimate the energy consumption (namely \texttt{CodeCarbon}), could have influenced the construct validity of our experiment.
To mitigate this threat, we ensured that the tool was made available as an open-source project (hence allowing us to independently scrutinize the appropriateness and correctness of the implementation), and that the tool relied on a widely used estimation method provided by a prominent technology company -- as we investigated, CodeCarbon uses the Intel\textsuperscript{\scriptsize\textregistered} Power Gadget\footnote{\url{https://www.intel.com/content/www/us/en/developer/articles/tool/power-gadget.html}. Accessed 28 January 2022.} tool under the hood. 

Regarding accuracy, we adopted a single metric, namely the F1-score. We chose this metrics over other ones, e.g., precision, recall, or logarithmic loss, as F1-score allowed us to provide a summary and intuitive presentation of the overall model accuracy~\textit{via} a single, well established, metric. To further mitigate potential threats related to the adoption of the F1-score, during the experiment execution, we also collected measurements of precision and recall, that are made available for scrutiny in the replication package of this study. 

\vspace{4pt}
\subsection{External Validity}

A prominent threat to the external validity of our study is posed by the adoption of a single experimental subject and a subset of AI algorithms. To mitigate this threat, we chose our experimental subject and independent variables to be as representative as possible. Specifically, we selected a common AI classification problem, namely text classification, selected a widely utilized peer-reviewed dataset~\cite{almeida2011contributions}, and considered a total of 6 different algorithms provided in the largely adopted Python library \texttt{scikit-learn}. 

As a further external validity threat, the experimental setup of this study did not integrate all the life cycle stages of AI models, as outlined by~\cite{kaack:hal-03368037}, and was limited to certain aspects of the~\textit{model training} phase. This outlines the \textit{in vitro} nature of our empirical experiment, which focused on studying data-centric approaches, and disregarded other aspects of AI model training (e.g., hyperparamether tuning) which would have commonly appeared in an \textit{in vivo} experimentation.
The narrow focus on energy consumption of data pre-processing of our study is intentional. More specifically, this research aimed at providing exploratory insights on the impact that AI data-centric approaches can have to AI energy consumption, and not how these approaches can be integrated in practice. As a result, numerous performance-optimization techniques common in a typical AI pipeline are excluded from our experiments, e.g., hyperparameter tuning, dimensionality reduction, and linear separability tests.


Albeit our best efforts, given the discussed threats to external validity, the results presented in this exploratory study have to be considered only as promising introductory insights, paving the way for future research on data-centric Green~AI.

\vspace{10pt}
\section{Conclusion and Future Work} \label{sec:conclusion}
With the popularization of large-scale datasets and affordable computational/storage capabilities, the energy consumed by AI is experiencing an unprecedented growth, which can no longer be neglected. With this study, we aim at exploring Green AI from a novel angle. Specifically, we investigate if modifying exclusively datasets, rather than the model training strategies, can optimize AI energy efficiency. To achieve our goal, we conduct an empirical experiment by considering 6 AI algorithms, two dataset modification strategies, and a dataset of over 5K~data~points. 

The results we obtained provide the first empirical evidence that not only data-centric strategies can be used to optimize AI energy efficiency, but also that such techniques can lead to a drastic energy consumption reduction. While AI accuracy may be negatively impacted by data-centric strategies, we also observed that in most cases such accuracy loss is negligible. 

From a practitioner perspective, our results highlight the high impact that dataset modifications have on AI energy efficiency, demonstrating that often ``designing for less'' while preprocessing a dataset can drastically reduce the energy consumed, while not sacrificing accuracy.

For researchers, our results open a new area of investigation, namely \textit{data-centric Green AI}, which, by considering the results documented in this research, demonstrates very~high~potential to address the sustainability of AI-based software-intensive systems. 

As future work, we plan to generalize our results by considering other AI application areas, e.g., image and audio recognition, and by utilizing additional large-scale datasets. Furthermore, we intend to investigate how other dataset modifications (e.g., data representations) may impact AI energy efficiency and related accuracy. Finally, we aim to conduct \textit{in vivo} experiments on data-centric Green AI, in order to investigate how data-centric techniques can be integrated in real-world AI pipelines, and how combining data-centric with other Green AI techniques (e.g., model training strategies) may impact energy efficiency and accuracy of AI models.
\clearpage
\bibliographystyle{IEEEtran}
\bibliography{biblio}

\begin{thebibliography}{10}
\providecommand{\url}[1]{#1}
\csname url@samestyle\endcsname
\providecommand{\newblock}{\relax}
\providecommand{\bibinfo}[2]{#2}
\providecommand{\BIBentrySTDinterwordspacing}{\spaceskip=0pt\relax}
\providecommand{\BIBentryALTinterwordstretchfactor}{4}
\providecommand{\BIBentryALTinterwordspacing}{\spaceskip=\fontdimen2\font plus
\BIBentryALTinterwordstretchfactor\fontdimen3\font minus
  \fontdimen4\font\relax}
\providecommand{\BIBforeignlanguage}[2]{{%
\expandafter\ifx\csname l@#1\endcsname\relax
\typeout{** WARNING: IEEEtran.bst: No hyphenation pattern has been}%
\typeout{** loaded for the language `#1'. Using the pattern for}%
\typeout{** the default language instead.}%
\else
\language=\csname l@#1\endcsname
\fi
#2}}
\providecommand{\BIBdecl}{\relax}
\BIBdecl

\bibitem{haakman2021ai}
M.~Haakman, L.~Cruz, H.~Huijgens, and A.~van Deursen, ``{AI} lifecycle models
  need to be revised,'' \emph{Empirical Software Engineering}, vol.~26, no.~5,
  pp. 1--29, 2021.

\bibitem{Strubell2019}
\BIBentryALTinterwordspacing
E.~Strubell, A.~Ganesh, and A.~McCallum, ``Energy and policy considerations for
  deep learning in nlp,'' 6 2019. [Online]. Available:
  \url{http://arxiv.org/abs/1906.02243}
\BIBentrySTDinterwordspacing

\bibitem{Lacoste2019}
\BIBentryALTinterwordspacing
A.~Lacoste, A.~Luccioni, V.~Schmidt, and T.~Dandres, ``Quantifying the carbon
  emissions of machine learning,'' 10 2019. [Online]. Available:
  \url{http://arxiv.org/abs/1910.09700}
\BIBentrySTDinterwordspacing

\bibitem{amodey2018ai}
\BIBentryALTinterwordspacing
D.~Amodei and D.~Hernandez, ``{AI} and compute,'' \emph{Open AI}, 2018.
  [Online]. Available: \url{https://openai.com/blog/ai-and-compute/}
\BIBentrySTDinterwordspacing

\bibitem{Bender2021}
E.~M. Bender, T.~Gebru, A.~McMillan-Major, and S.~Shmitchell, ``On the dangers
  of stochastic parrots: Can language models be too big?''\hskip 1em plus 0.5em
  minus 0.4em\relax Association for Computing Machinery, Inc, 3 2021, pp.
  610--623.

\bibitem{Schwartz2020}
R.~Schwartz, J.~Dodge, N.~A. Smith, and O.~Etzioni, ``Green {AI},''
  \emph{Communications of the ACM}, vol.~63, pp. 54--63, 11 2020.

\bibitem{calero2021investigating}
C.~Calero, M.~Polo, and M.~{\'A}. Moraga, ``Investigating the impact on
  execution time and energy consumption of developing with spring,''
  \emph{Sustainable Computing: Informatics and Systems}, vol.~32, p. 100603,
  2021.

\bibitem{oliveira2019recommending}
W.~Oliveira, R.~Oliveira, F.~Castor, B.~Fernandes, and G.~Pinto, ``Recommending
  energy-efficient java collections,'' in \emph{2019 IEEE/ACM 16th
  International Conference on Mining Software Repositories (MSR)}.\hskip 1em
  plus 0.5em minus 0.4em\relax IEEE, 2019, pp. 160--170.

\bibitem{pereira2021ranking}
R.~Pereira, M.~Couto, F.~Ribeiro, R.~Rua, J.~Cunha, J.~P. Fernandes, and
  J.~Saraiva, ``Ranking programming languages by energy efficiency,''
  \emph{Science of Computer Programming}, vol. 205, p. 102609, 2021.

\bibitem{georgiou2018your}
S.~Georgiou, M.~Kechagia, P.~Louridas, and D.~Spinellis, ``What are your
  programming language's energy-delay implications?'' in \emph{Proceedings of
  the 15th International Conference on Mining Software Repositories}, 2018, pp.
  303--313.

\bibitem{anwar2020should}
H.~Anwar, B.~Demirer, D.~Pfahl, and S.~Srirama, ``Should energy consumption
  influence the choice of android third-party http libraries?'' in
  \emph{Proceedings of the IEEE/ACM 7th International Conference on Mobile
  Software Engineering and Systems}, 2020, pp. 87--97.

\bibitem{cruz2019catalog}
L.~Cruz and R.~Abreu, ``Catalog of energy patterns for mobile applications,''
  \emph{Empirical Software Engineering}, vol.~24, no.~4, pp. 2209--2235, 2019.

\bibitem{cruz2019do}
L.~Cruz, R.~Abreu, J.~Grundy, L.~Li, and X.~Xia, ``Do energy-oriented changes
  hinder maintainability?'' in \emph{2019 IEEE International Conference on
  Software Maintenance and Evolution (ICSME)}, 2019, pp. 29--40.

\bibitem{venters2018software}
C.~C. Venters, R.~Capilla, S.~Betz, B.~Penzenstadler, T.~Crick, S.~Crouch,
  E.~Y. Nakagawa, C.~Becker, and C.~Carrillo, ``Software sustainability:
  Research and practice from a software architecture viewpoint,'' \emph{Journal
  of Systems and Software}, vol. 138, pp. 174--188, 2018.

\bibitem{verdecchia2018empirical}
R.~Verdecchia, R.~A. Saez, G.~Procaccianti, and P.~Lago, ``Empirical evaluation
  of the energy impact of refactoring code smells.'' in \emph{ICT4S}, 2018, pp.
  365--383.

\bibitem{verdecchia2017estimating}
R.~Verdecchia, G.~Procaccianti, I.~Malavolta, P.~Lago, and J.~Koedijk,
  ``Estimating energy impact of software releases and deployment strategies:
  The kpmg case study,'' in \emph{2017 ACM/IEEE International Symposium on
  Empirical Software Engineering and Measurement (ESEM)}.\hskip 1em plus 0.5em
  minus 0.4em\relax IEEE, 2017, pp. 257--266.

\bibitem{ribeiro2021ecoandroid}
A.~Ribeiro, J.~F. Ferreira, and A.~Mendes, ``Ecoandroid: An android studio
  plugin for developing energy-efficient java mobile applications,'' 2021.

\bibitem{cruz2019emaas}
L.~Cruz and R.~Abreu, ``Emaas: Energy measurements as a service for mobile
  applications,'' in \emph{2019 IEEE/ACM 41st International Conference on
  Software Engineering: New Ideas and Emerging Results (ICSE-NIER)}, 2019, pp.
  101--104.

\bibitem{wyn}
A.~van Wynsberghe, ``{Sustainable {AI}: {AI} for sustainability and the
  sustainability of AI},'' \emph{AI and Ethics}, pp. 1--6, 2021.

\bibitem{kloh}
V.~Klôh, B.~Schulze, and M.~Ferro, ``Use of machine learning for improvements
  in performance and energy consumption in hpc systems,'' 09 2020.

\bibitem{zhu}
S.~Zhu, K.~Ota, and M.~Dong, ``Green {AI} for {IIoT}: Energy efficient
  intelligent edge computing for industrial internet of things,'' \emph{IEEE
  Transactions on Green Communications and Networking}, 08 2021.

\bibitem{Wu2021}
\BIBentryALTinterwordspacing
C.-J. Wu, R.~Raghavendra, U.~Gupta, B.~Acun, N.~Ardalani, K.~Maeng, G.~Chang,
  F.~A. Behram, J.~Huang, C.~Bai, M.~Gschwind, A.~Gupta, M.~Ott, A.~Melnikov,
  S.~Candido, D.~Brooks, G.~Chauhan, B.~Lee, H.-H.~S. Lee, B.~Akyildiz,
  M.~Balandat, J.~Spisak, R.~Jain, M.~Rabbat, and K.~Hazelwood, ``Sustainable
  {AI}: Environmental implications, challenges and opportunities,'' 10 2021.
  [Online]. Available: \url{http://arxiv.org/abs/2111.00364}
\BIBentrySTDinterwordspacing

\bibitem{yang2021tuning}
G.~Yang, E.~Hu, I.~Babuschkin, S.~Sidor, X.~Liu, D.~Farhi, N.~Ryder,
  J.~Pachocki, W.~Chen, and J.~Gao, ``Tuning large neural networks via
  zero-shot hyperparameter transfer,'' \emph{Advances in Neural Information
  Processing Systems}, vol.~34, 2021.

\bibitem{zogaj2021doing}
F.~Zogaj, J.~P. Cambronero, M.~C. Rinard, and J.~Cito, ``Doing more with less:
  characterizing dataset downsampling for automl,'' \emph{Proceedings of the
  VLDB Endowment}, vol.~14, no.~11, pp. 2059--2072, 2021.

\bibitem{martin}
E.~Garcia-Martin, N.~Lavesson, and H.~Grahn, ``Identification of energy
  hotspots: A case study of the very fast decision tree,'' \emph{Green,
  Pervasive, and Cloud Computing}, pp. 267--281, 01 2017.

\bibitem{mcintosh2018MLEnergy}
A.~McIntosh, S.~Hassan, and A.~Hindle, ``What can android mobile app developers
  do about the energy consumption of machine learning?'' \emph{Empirical
  Software Engineering}, pp. 1--42, May 2018.

\bibitem{georgiou2022green}
S.~Georgiou, M.~Kechagia, T.~Sharma, F.~Sarro, and Y.~Zou, ``{Green AI: Do Deep
  Learning Frameworks Have Different Costs?}'' in \emph{2022 IEEE/ACM 44st
  International Conference on Software Engineering (ICSE)}, 2022.

\bibitem{gqm}
V.~R. Basili, G.~Caldiera, and D.~Rombach, ``The {G}oal {Q}uestion {M}etric
  {A}pproach,'' in \emph{Encyclopedia of {S}oftware {E}ngineering}.\hskip 1em
  plus 0.5em minus 0.4em\relax Wiley, 1994, pp. 528--532.

\bibitem{almeida2011contributions}
T.~A. Almeida, J.~M.~G. Hidalgo, and A.~Yamakami, ``Contributions to the study
  of sms spam filtering: new collection and results,'' in \emph{Proceedings of
  the 11th ACM symposium on Document engineering}, 2011, pp. 259--262.

\bibitem{liu1995chi2}
H.~Liu and R.~Setiono, ``Chi2: Feature selection and discretization of numeric
  attributes,'' in \emph{Proceedings of 7th IEEE International Conference on
  Tools with Artificial Intelligence}.\hskip 1em plus 0.5em minus 0.4em\relax
  IEEE, 1995, pp. 388--391.

\bibitem{radovanovic2021carbonaware}
A.~Radovanovic, R.~Koningstein, I.~Schneider, B.~Chen, A.~Duarte, B.~Roy,
  D.~Xiao, M.~Haridasan, P.~Hung, N.~Care, S.~Talukdar, E.~Mullen, K.~Smith,
  M.~Cottman, and W.~Cirne, ``Carbon-aware computing for datacenters,'' 2021.

\bibitem{garciamartin}
E.~García-Martín, C.~F. Rodrigues, G.~Riley, and H.~Grahn, ``Estimation of
  energy consumption in machine learning,'' \emph{Journal of Parallel and
  Distributed Computing}, vol. 134, pp. 75--88, 12 2019.

\bibitem{nguyen2021dataset}
T.~Nguyen, Z.~Chen, and J.~Lee, ``Dataset meta-learning from kernel
  ridge-regression,'' in \emph{International Conference on Learning
  Representations}, 2021.

\bibitem{zhao2021dataset}
B.~Zhao and H.~Bilen, ``Dataset condensation with differentiable siamese
  augmentation,'' in \emph{Proceedings of the 38th International Conference on
  Machine Learning}, ser. Proceedings of Machine Learning Research, M.~Meila
  and T.~Zhang, Eds., vol. 139.\hskip 1em plus 0.5em minus 0.4em\relax PMLR,
  18--24 Jul 2021, pp. 12\,674--12\,685.

\bibitem{borsos2020coresets}
Z.~Borsos, M.~Mutny, and A.~Krause, ``Coresets via bilevel optimization for
  continual learning and streaming,'' in \emph{Advances in Neural Information
  Processing Systems}, H.~Larochelle, M.~Ranzato, R.~Hadsell, M.~F. Balcan, and
  H.~Lin, Eds., vol.~33.\hskip 1em plus 0.5em minus 0.4em\relax Curran
  Associates, Inc., 2020, pp. 14\,879--14\,890.

\bibitem{verdecchia2021green}
R.~Verdecchia, P.~Lago, C.~Ebert, and C.~De~Vries, ``Green {IT} and green
  software,'' \emph{IEEE Software}, vol.~38, no.~6, pp. 7--15, 2021.

\bibitem{alcott2005jevons}
B.~Alcott, ``Jevons' paradox,'' \emph{Ecological economics}, vol.~54, no.~1,
  pp. 9--21, 2005.

\bibitem{wohlin2012experimentation}
C.~Wohlin, P.~Runeson, M.~H{\"o}st, M.~C. Ohlsson, B.~Regnell, and
  A.~Wessl{\'e}n, \emph{Experimentation in software engineering}.\hskip 1em
  plus 0.5em minus 0.4em\relax Springer Science \& Business Media, 2012.

\bibitem{hindle2014greenminer}
A.~Hindle, A.~Wilson, K.~Rasmussen, E.~J. Barlow, J.~C. Campbell, and
  S.~Romansky, ``Greenminer: A hardware based mining software repositories
  software energy consumption framework,'' in \emph{Proceedings of the 11th
  working conference on mining software repositories}, 2014, pp. 12--21.

\bibitem{verdecchia2018code}
R.~Verdecchia, A.~Guldner, Y.~Becker, and E.~Kern, ``Code-level energy hotspot
  localization via naive spectrum based testing,'' in \emph{Advances and New
  Trends in Environmental Informatics}.\hskip 1em plus 0.5em minus 0.4em\relax
  Springer, 2018, pp. 111--130.

\bibitem{feitosa2017investigating}
D.~Feitosa, R.~Alders, A.~Ampatzoglou, P.~Avgeriou, and E.~Y. Nakagawa,
  ``Investigating the effect of design patterns on energy consumption,''
  \emph{Journal of Software: Evolution and Process}, vol.~29, no.~2, p. e1851,
  2017.

\bibitem{kaack:hal-03368037}
\BIBentryALTinterwordspacing
L.~H. Kaack, P.~L. Donti, E.~Strubell, G.~Kamiya, F.~Creutzig, and D.~Rolnick,
  ``{Aligning artificial intelligence with climate change mitigation},'' Oct.
  2021, working paper or preprint. [Online]. Available:
  \url{https://hal.archives-ouvertes.fr/hal-03368037}
\BIBentrySTDinterwordspacing

\end{thebibliography}

\end{document}